\documentclass{article}
\usepackage{ijcai24}

\usepackage{times}
\usepackage{soul}
\usepackage{url}
\usepackage[hidelinks]{hyperref}
\usepackage[utf8]{inputenc}
\usepackage[small]{caption}
\usepackage{graphicx}
\usepackage{amsmath}
\usepackage{amsthm}
\usepackage{amssymb}
\usepackage{bm}
\usepackage{dsfont}
\usepackage{booktabs}
\usepackage{algorithm}
\usepackage{algorithmic}
\usepackage[switch]{lineno}
\usepackage{todonotes}
\usepackage{xcolor}
\usepackage[inline]{enumitem}
\usepackage{multirow}
\usepackage{tabularx}
\usepackage{subfig}
\usepackage{makecell}

\newtheorem{definition}{Definition}
\newtheorem{corollary}{Corollary}

\newtheorem{theorem}{Theorem}

\title{Sandra - A Neuro-Symbolic Reasoner Based On Descriptions And Situations}

\author{
Nicolas Lazzari$^1$
\and
Stefano De Giorgis$^1$\and
Aldo Gangemi $^1$\And
Valentina Presutti$^1$
\affiliations
$^1$University of Bologna
\emails
\{nicolas.lazzari3, stefano.degiorgis2, aldo.gangemi, valentina.presutti\}@unibo.it
}

\begin{document}

\maketitle

\begin{abstract}
This paper presents \emph{sandra}, a neuro-symbolic reasoner combining vectorial representations with deductive reasoning. \emph{Sandra} builds a vector space constrained by an ontology and performs reasoning over it. The geometric nature of the reasoner allows its combination with neural networks, bridging the gap with symbolic knowledge representations. \emph{Sandra} is based on the Description and Situation (DnS) ontology design pattern, a formalization of frame semantics. Given a set of facts (a situation) it allows to infer all possible perspectives (descriptions) that can provide a plausible interpretation for it, even in presence of incomplete information. We prove that our method is correct with respect to the DnS model. We experiment with two different tasks and their standard benchmarks, demonstrating that, without increasing complexity, \emph{sandra} (i) outperforms all the baselines (ii) provides interpretability in the classification process, and (iii) allows control over the vector space, which is designed a priori. 
\end{abstract}

\section{Introduction} 
\label{sec:intro}

Reasoning on perspectives is relevant in many contexts and applications. In very rigorous domains such as medicine or jurisprudence, the same case can be interpreted through more than one description: a patient's situation can be analyzed through different diagnostic hypotheses; a legal state of affairs can be interpreted by applying different, possibly conflicting, norms. In storytelling techniques, spin doctors can frame the same topic in different ways to support alternative political scenarios: ``\emph{Conservatives claim that we need relief from taxes vs. Democrats claim that taxes are investments for us.}'' \cite{vossen2022perspective}. 

\begin{figure}[ht]
    \centering
    \setlength{\belowcaptionskip}{-10pt}
    \includegraphics[width=.6\columnwidth]{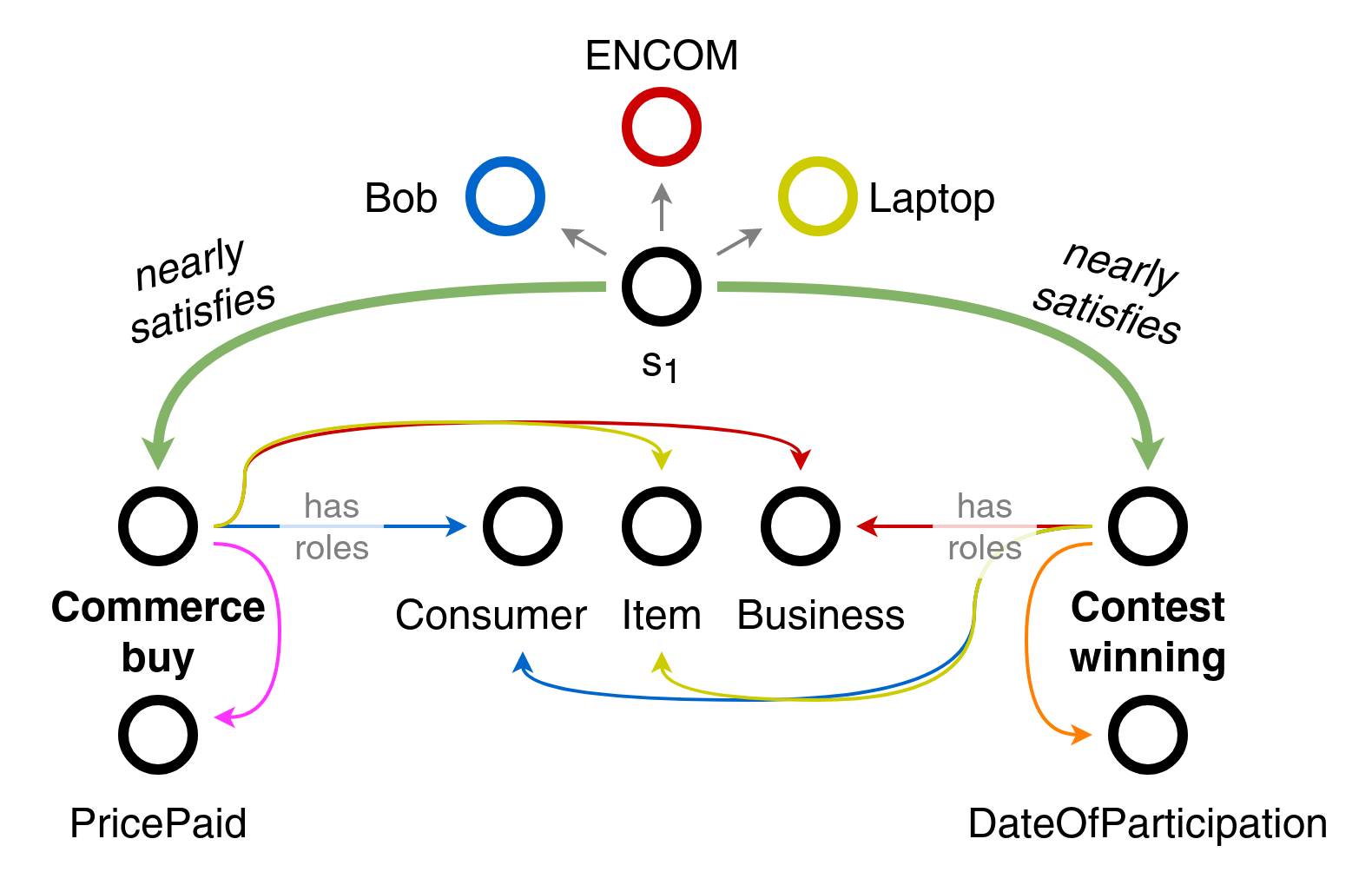}
    \caption{Example of two descriptions (\texttt{Commerce buy} and \texttt{Contest winning}) that are satisfied by a situation that involves the entities \texttt{Bob}, \texttt{ENCOM}, \texttt{Laptop}. The two descriptions define the same roles, hence they provide two different perspectives from which the situation can be interpreted.}
    \label{fig:DnS}
\end{figure}

Informally, a set of facts (situation) can be described through more than one - potentially conflicting - perspective, which can be inferred even in presence of  partial data. This paper proposes a method and a neuro-symbolic reasoner, named \emph{sandra}, able to infer all perspectives that are plausible descriptions for a given situation. 
\begin{figure*}[h]
    \centering
    \includegraphics[width=\textwidth]{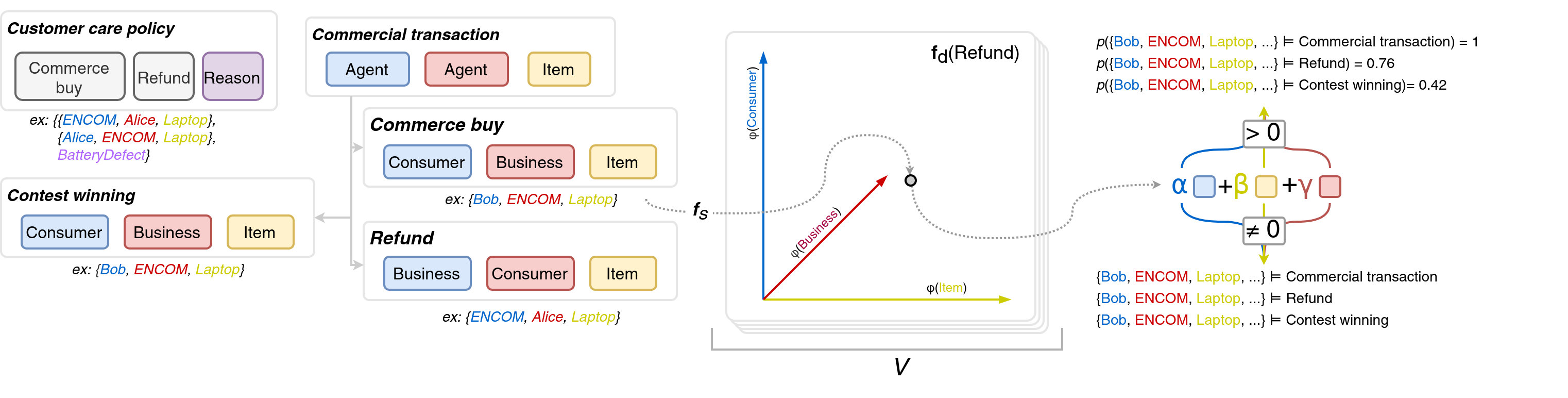}
    \caption{Examples of some descriptions and situations alongside the conversion of a situation into $V$ to detect the satisfied descriptions. The process of deducing which descriptions are satisfied by $\bm{f}_s$ is shown on the right.}
    \label{fig:sandra-diagram}
\end{figure*}
Associating different plausible descriptions to a situation requires the capability to reason at both the intensional and the extensional level, i.e., with both concepts and facts, 
posing a challenge on how to formally represent the knowledge required, as well as on the reasoning technique to apply. 

From a knowledge representation perspective, the Description and Situation (DnS) \cite{gangemi2003understanding} ontology design pattern addresses this problem. DnS provides a formalization of frame semantics, generalizing over Fillmore's \cite{fillmore2006frame} and Minsky's \cite{minsky1974framework} proposals. It has been largely used in many ontology projects (mainly using its OWL formalization) \cite{10.1145/1597735.1597760,723935,Porzel2020WhatSY,10.3233/SW-200416}. DnS defines a general vocabulary for n-ary relations, introducing the concepts of \emph{description} and \emph{situation}. A situation is a set of facts, as it is described by an observer, involving a set of entities.
In Figure \ref{fig:DnS} a situation involving three entities is represented: \texttt{Bob}, \texttt{ENCOM}, and \texttt{Laptop}. A description is a perspective (a theory, a schema) defining concepts that can classify, hence interpret, the entities observed in a situation. In Figure \ref{fig:DnS} two descriptions are represented, \texttt{Commerce buy} and \texttt{Contest winning}. With DnS, both descriptions and situations are formalized as first-order entities in an ontology, therefore potentially enabling reasoning over them both at the intensional (description) and extensional (situation) level, in the same domain of discourse. In DnS terms, we say that a situation $s$ \emph{satisfies} a description $d$, when $d$ is a plausible or correct interpretation of $s$. In Figure \ref{fig:DnS} both descriptions (\texttt{Commerce buy} and \texttt{Contest winning}) are satisfied by the situation $s_1$ involving \texttt{Bob}, \texttt{ENCOM}, and \texttt{Laptop}, even if they are conceptually incompatible. A perspective-based reasoner shall be able to detect this conceptual difference and yet infer both \texttt{satisfies} relations.

To the best of our knowledge, this behaviour is unobtainable with existing logic-based reasoners. Within a traditional logic approach, conflicting concepts (such as \texttt{Commerce buy} and \texttt{Contest winning}) are correctly modelled as disjoint classes. Therefore, classifying the situation from the example of Figure \ref{fig:DnS} in both descriptions would lead to an inconsistency. Furthermore, no classification would be provided if the extensional data available is partial. Fuzzy reasoners, such as \cite{kamide2022fuzzydl}, may be more tolerant but would show the same problem (inconsistency) adding an issue of possible undecidability. Logic-based representations and reasoning have also benefits: a clear, formal model of descriptions is defined and the inference is always sound and complete with respect to that model. At the same time, the data representing situations must be structured and compliant with such model, requiring significant effort. 

Neural Network models are a solution to both obtaining an approximate classification, and being able to inject any type of data (images, text, etc.), not only structured data. Nevertheless, they may need a large sample including both descriptions and situations, and the results may be incompatible with the formal model, for example it might happen that only some of the plausible descriptions are retrieved. In short, logic reasoning lacks inconsistency and partial-data-tolerant deductions, while neural network models lack the completeness of deductive inference. 

Neuro-symbolic methods, such as \cite{manhaeve2018deepproblog,geh2023dpasp,van2022nesi}, combine important benefits of logic-based and approximate reasoning, and (very important) they can ingest unstructured data. Nevertheless, they rely on the same consistency assumption, which is undesirable in a perspective-reasoning scenario. 

The approach proposed in this research (\emph{sandra}) relies on the neuro-symbolic paradigm but addresses this issue. It formalizes and implements the inference of the \texttt{satisfies} relation between situations and descriptions and, at the same time, it provides a probability score and a deductive, interpretable inference. 

It includes:
(i) a theoretical framework formalizing DnS in a differentiable and probabilistic fashion, which we prove to be correct with respect to DnS. It creates a vector space $V$ isomorphic to a DnS-based ontology (defining a set of descriptions); (ii) the implementation of a neuro-symbolic architecture, where the neural network is constrained to position any detected situation (from any source type) in one or more subspaces of $V$, each corresponding to a description defined in the ontology. The intuition is that this constraint makes the network less prone to noise. Our formulation allows efficient inference in the DnS domain. 
Our hypothesis is that by constraining the representation learning process to a DnS-based ontology we can perform perspective-based reasoning without performance loss or increase of computational complexity.

To evaluate our hypothesis we experiment our method on two different tasks: visual reasoning on the I-RAVEN benchmark \cite{hu2021stratified} and domain generalization on Fashion-MNIST \cite{xiao2017fashionmnist}. For each task, we asses the influence of adding \emph{sandra} to a standard baseline model and find increased performances with polynomial complexity~\footnote{In time and space with respect to the number of entities within an ontology.}.


Our contribution can be summarized as follows:
\begin{enumerate*}
    \item A correct differentiable probabilistic formalization of DnS;
    \item A neuro-symbolic reasoner that combines deductive and inductive reasoning to classify arbitrary situations extracted by a neural network into the descriptions (perspectives) that can plausibly interpret them. 
\end{enumerate*}

The rest of the paper provides a formalization of DnS (Section \ref{sec:dns}) and of our method (Section \ref{sec:method}). Section \ref{sec:experiments} describes our experiments and results, which are further discussed, along with future work in Section \ref{sec:discussion}. Section \ref{sec:related} discusses relevant related work, while Section \ref{sec:conclusion} concludes the paper.

\section{Descriptions \& Situations} 
\label{sec:dns}

In DnS, descriptions and situations are n-ary relations. We refer to the arguments of a description as \emph{roles}, and to those of a situation as \emph{entities}. For example, in Figure \ref{fig:DnS}, \texttt{Business}, \texttt{Consumer}, and \texttt{Item} are the roles of the \texttt{Commerce buy} description, while \texttt{ENCOM}, \texttt{Bob}, and \texttt{Laptop} are the entities of the situation $s_1$. A situation $s$ \texttt{satisfies} a description $d$, if each entity of $s$ can be classified by one role of $d$~\footnote{In the original formalization of DnS this is called maximal satisfaction. We capture all possible types of satisfaction although the terminology is simplified.}. 


A formalization of DnS in First Order Logic is given in \cite{gangemi2003understanding}. We extend it to allow roles as n-ary relations and provide here the resulting formal semantics.
With $\mathcal{O}$ a DnS-based ontology, we define $\mathcal{D}$ as the set of descriptions in the ontology, with $\mathcal{R}$ the set of roles and $\mathcal{E}$ the set of entities and assume that $\mathcal{D} \cup \mathcal{R} \cup \mathcal{E} \subset \mathcal{O}$.

\begin{definition}
    A \underline{description} $d$ is a set $d = \{ r_1, \cdots, r_n \} \in \mathcal{D}$ with $r_i \in \mathcal{R} \cup (\mathcal{D} - \{ d \})$.
    \label{def:dns-description}
\end{definition}

In Figure \ref{fig:sandra-diagram}, $\textit{Refund} = \{ \textit{Consumer}, \textit{Business}, \textit{Item}\} \in \mathcal{D}$.

Notice that descriptions can be used as roles as well. For instance, $\textit{Customer care policy} = \{ \textit{Commerce buy}, \textit{Refund}, \textit{Reason} \} \in \mathcal{D}$, where $\textit{Commerce buy}, \textit{Refund} \in \mathcal{D}$.

\begin{definition}
    A \underline{situation} $s$ is a set $s = (e_1, \cdots, e_n) \in \mathcal{S}$ with $e_i \in \mathcal{D} \cup \mathcal{R} \cup \mathcal{E} \cup \mathcal{S}$.
    \label{def:dns-situation}
\end{definition}

For example, the set $\{ \textit{Bob}, \textit{ENCOM}, \textit{Laptop} \}$ of Figure \ref{fig:sandra-diagram} is a situation. 

A situation can also involve other situations as its entities. For example, in Figure \ref{fig:sandra-diagram}, the set
\begin{align*}
 s = \{ & \{ \textit{ENCOM}, \textit{Alice}, \textit{Laptop} \}, \\
        & \{ \textit{Alice}, \textit{ENCOM}, \textit{Laptop}\}, \\
        & \{ \textit{BatteryDefect} \} \}
\end{align*}

represents a situation in which \textit{Alice} bought a \textit{Laptop} from \textit{ENCOM}, \textit{ENCOM} received it back from \textit{Alice} and a \textit{BatteryDefect} reason is specified. Both \textit{Alice} buying and \textit{ENCOM} receiving the \textit{Laptop} are two separate situations that are part of another situation, which also involves \textit{BatteryDefect}.


When a description $d$ explains a situation $s$, we say that $s$ \textit{satisfies} $d$, written as $d \models s$.

\begin{definition}[satisfaction] \label{def:dns-satisfies-definition}
    With $d \in \mathcal{D}$ and $s \in \mathcal{S}$
    \begin{equation*}
        d \models s \Leftrightarrow \forall r \in d. (\exists e \in  s. e \in r) \lor (\exists s' \in s . r \models s')   
    \end{equation*}
\end{definition}


In Figure \ref{fig:sandra-diagram} the situation $s = \{ \textit{Bob}, \textit{ENCOM}, \textit{Laptop} \}$ satisfies the descriptions \textit{Commerce buy} and \textit{Contest winning}. Both descriptions are perspectives from which $s$ can be interpreted. 


Assuming that $\mathcal{O}$ asserts that $\textit{Consumer}, \textit{Business} \subseteq \textit{Agent}$ (i.e. \textit{Agent} is a broader role than \textit{Consumer} and \textit{Business}), we have that $\textit{Commercial transaction} \models s$ as well, since the two roles are also \textit{Agents}.

\begin{corollary}
    Given $d, d' \in \mathcal{D}$ with $d' \subseteq d$ and $s \in \mathcal{S}$ then $d' \models s \Rightarrow d \models s$.
    \label{cor:dns-satisfies-superclass}
\end{corollary}

Corollary \ref{cor:dns-satisfies-superclass} trivially follows from definition \ref{def:dns-satisfies-definition}.

Definition \ref{def:dns-satisfies-definition} states the condition such that a perspective is valid for a situation. However, a description can be a plausible perspective for a situation even if only some of its roles are fulfilled. In this case, we say that a description \emph{nearly} satisfies a situation, written as $d \widetilde{\models} s$.

\begin{definition}[\textit{near}-satisfaction] \label{def:dns-nearly-satisfies-definition}
    With $d \in \mathcal{D}$ and $s \in \mathcal{S}$,
    \begin{equation*}   
        d \widetilde{\models} s \Leftrightarrow \exists r \in d. (\exists e \in  s. e \in r) \lor (\exists s' \in s . r \models s')
    \end{equation*}
\end{definition}

From Definition \ref{def:dns-nearly-satisfies-definition} it follows that if a situation satisfies a description, then it also \textit{nearly}-satisfies it.

\begin{corollary} \label{def:satisfies-implies-nearsatisfies}
    With $d \in \mathcal{D}$ and $s \in \mathcal{S}$, $d \models s \Rightarrow d \widetilde{\models} s$
\end{corollary}

The corollary trivially follows from Definitions \ref{def:dns-satisfies-definition} and \ref{def:dns-nearly-satisfies-definition}.

\section{Method}
\label{sec:method}
With reference to Section \ref{sec:dns} we define a vector space $V$ over the field $\mathbb{R}$ with $dim(V) = |\mathcal{R} \cup \mathcal{D}|$. For each description $d \in \mathcal{D}$ a subspace $V_d$ is identified by its roles. Assuming that $\mathbf{v}_s$ is a vector representing $s \in \mathcal{S}$ and that $d \models s$ with $d \in \mathcal{D}$, the intuition following from Definition \ref{def:dns-satisfies-definition} (and \ref{def:dns-nearly-satisfies-definition}) is that $\mathbf{v}_s$ must be defined by (some) entities classified by the roles in $\mathcal{D}$ that identify the subspace $V_d$.

In this section, we describe how a subspace $V_d$ is defined and how to deduce when a description is (\textit{nearly-})satisfied by a situation.
Notation-wise, we assume that $s \in \mathcal{S}$ and for some $d \in \mathcal{D}$, $V_d$ is the subspace associated with $d$ and $B_d$ is a spanning set of $V_d$.

\paragraph{Subspace definition}
The set $B_d$, defined as $B_d = \{\mathbf{b}_1, \cdots, \mathbf{b}_n \}$ is a basis for the subspace $V_d$, i.e. the vectors in $B_d$ form a minimal spanning set of $V_d$. Hence every $\mathbf{v} \in V_d$ can be expressed as a linear combination (a weighted sum) of the vectors in $B_d$ \cite{meyer2023linearalgebra}. 

Given Definitions \ref{def:dns-satisfies-definition} and \ref{def:dns-nearly-satisfies-definition}, it follows that the vectors in $B_d$ must originate from the description $d$ and its roles $r \in d$. We define $\bm{f}_d$ as the function that converts each description $d \in \mathcal{D}$ to a vector $\bm{f}_d(d) \in B_d$. Informally, $\bm{f}_d(d)$ is obtained by recursively aggregating the vector representations $\bm{v}_r$ of the roles $r \in d$. A function $\phi$ computes $\bm{v}_r$ ensuring that $\bm{f}_d(d) \in B_d$ and that $\bm{v}_r$ reflects subsumption relations between roles. 


\begin{definition}[$\bm{f}_d$] \label{def:fd}
Given $x \in \mathcal{D} \cup \mathcal{R}$, $\bm{f}_d: \mathcal{D} \cup \mathcal{R} \rightarrow V$ is defined as 
$$ \bm{f}_d(x) = \phi(x) + \mathds{1}_{[x \in \mathcal{D}]} \sum_{r \in x} \bm{f}_d(r) $$
where $\phi(x) = [ \mathds{1}_{[x \subseteq y_1]} || \cdots || \mathds{1}_{[x \subseteq y_n]}]$ with $y_1, \cdots, y_n \in \mathcal{R} \cup \mathcal{D}$ and $n = |\mathcal{D} \cup \mathcal{R}|$.
\end{definition}

The function $\phi(x)$ considers $\mathcal{D} \cup \mathcal{R}$ as an ordered set with $n = |\mathcal{D} \cup \mathcal{R}|$ elements. It creates a vector of dimension $n$ where the $i_{th}$ element is 1 if $x$ is equal or a subset of the $i_{th}$ element in $\mathcal{D} \cup \mathcal{R}$.
For example, consider $\mathcal{D} \cup \mathcal{R} = \{ \texttt{Agent}, \texttt{Item}, \texttt{Consumer}, \cdots \}$ with $|\mathcal{D} \cup \mathcal{R}| = 10$ and $\texttt{Consumer} \subseteq \texttt{Agent}$ as in Figure \ref{fig:sandra-diagram}. Then $\bm{f}_d(\texttt{Agent}) = \phi(a) = [1, 0, 0, 0, 0, 0, 0, 0, 0, 0]$, $\bm{f}_d(\texttt{Consumer}) = \phi(d) = [1, 0, 1, 0, 0, 0, 0, 0, 0, 0]$ and $\bm{f}_d(\texttt{Commercial transaction}) = \phi(\texttt{Agent}) + \phi(\texttt{Item})  = [1, 1, 0, 0, 0, 0, 0, 0, 0, 0]$.

\begin{theorem} \label{theo:subspace-basis}
    Given a description $d \in \mathcal{D}$, the vectors in $B = \{\bm{f}_d(r_1), \cdots, \bm{f}_d(r_n)\}$ with $r_1, \cdots, r_n \in d$ are linearly independent.
\end{theorem}


\begin{proof}
Without loss of generality, assume that $\mathcal{O}$ is converted to a Directed Acyclic Graph (DAG) where edges connect descriptions to their roles. Consider $\alpha_1 \bm{f}_d(x_1) + \cdots + \alpha_n \bm{f}_d(x_n)$ with $x_1 \neq \cdots \neq x_n \in \mathcal{R} \cup \mathcal{D}$ and $\alpha_1, \cdots, \alpha_n \in \mathds{R}$. Each term $\alpha_i \bm{f}_d(x_i) = \sum_{r \in X_i} \alpha_i \phi(r)$ where $X_i$ is the union of $x_i \setminus \mathcal{D}$ and of the roles recursively collected from the nested descriptions in $x_i$. Since $\phi$ is positive-definite and injective ($\exists \hat{x} . \phi(\hat{x}) = a \neq b = \phi(\hat{x}) \rightarrow \exists \hat{y}. \mathds{1}_{[\hat{x} \subseteq \hat{y} ]} \neq \mathds{1}_{[\hat{x} \subseteq \hat{y} ]} \rightarrow \bot$) then $\alpha_1 \bm{f}_d(x_1) + \cdots + \alpha_n \bm{f}_d(x_n) = 0 \Leftrightarrow \alpha_1 = \cdots = \alpha_n = 0$. Thus, $\bm{f}_d(x_1), \cdots, \bm{f}_d(x_n)$ are linearly independent.
\end{proof}

By taking $B_d$ as the spanning set of $V_d$, it forms a basis of the subspace associated with the description $d$. It will be possible to represent a situation $s$ that satisfy $d$ as a linear combination of the vectors in $B_d$.

\begin{definition}[$B_d$ and $V_d$]
    Given a description $d \in \mathcal{D}$, $V$ has a subspace $V_d$ spanned by the basis $B_d = \{ \bm{f}_d(r_1), \cdots, \bm{f}_d(r_n) \}$ with $r_1, \cdots, r_n \in d$.
\end{definition}

It follows from Definition \ref{def:fd} that for any $d \in \mathcal{D}$, $B_d$ can be computed in polynomial time $\mathcal{O}(\mathcal{|D}|^2)$. The proof follows from Definition \ref{def:fd}: by converting $\mathcal{O}$ to a DAG, for each description we have to loop through every other description in the worst case. 

\paragraph{Situation encoding}
Given the definition of the subspace $V_d$, the intuition is that its vectors are those representing all the situations $s$ such that $d \models s$. 
To obtain $\bm{v}_s \in V$, the vector representation of $s$, we introduce the function $\bm{f}_s$, which maps a situation $s \in \mathcal{S}$ into a vector.

\begin{definition} \label{def:fs}
    Given $s \in \mathcal{S}$, $e \in \mathcal{E}$ and $r \in \mathcal{R}$, and with $\psi: \mathcal{E} \to \mathcal{R}$ defined such that $e \in r \Rightarrow \psi(e) = r$, the function $\bm{f}_s: \mathcal{S} \rightarrow V$ is defined as
    \begin{equation*}
       \bm{f}_s(s) = \sum\limits_{e \in s} \begin{cases}
           \bm{f}_s(e) & \text{if } e \in \mathcal{S} \\
           \bm{f}_d(\psi(e)) & \text{otherwise}
       \end{cases}
    \end{equation*}
\end{definition}

The function $\psi$ can be implemented by a deductive reasoner (e.g. \cite{glimm2014hermit}) or provided externally. 

The vector $\bm{v}_s$ produced by the function $\bm{f}_s$ is the aggregation of the vector representation of the roles that classifies the entities in $s$ and its nested situations. Nested situations are recursively converted into a vector using the same function $\bm{f}_s$. For example from Figure \ref{fig:sandra-diagram}, $\bm{f}_s(\{ \texttt{Bob}, \texttt{ENCOM}, \texttt{Laptop} \}) = \bm{f}_d(\texttt{Consumer}) + \bm{f}_d(\texttt{Business}) + \bm{f}_d(\texttt{Item})$. Note that from Definition \ref{def:fd} it follows that $\bm{f}_s$ is a positive definite function, i.e. $\bm{f}_s(x) \geq 0 \ \forall s \in \mathcal{S}$, since $\phi$ is positive definite.

\paragraph{Description satisfaction} 
Given Theorem \ref{theo:subspace-basis} and function $\bm{f}_s$, we can check whether the vector representation $\bm{v}_s$ lies in the subspace $V_d$ by checking whether $\bm{v}_s$ is linear combination of the vectors in $B_d$.

Consider $A_d$ as the $|\mathcal{R} \cup \mathcal{D}| \times |d|$ matrix whose row-space is $B_d$ (i.e. the rows are formed by the vectors $\bm{b} \in B_d$). 
The solution $\bm{x}$ to the linear system $A_d \bm{x} = \bm{v}_s$ contains the coefficients to obtain the vector $\bm{v}_s$ as the linear combination of the vectors $\bm{b} \in B_d$. With $A_d^+$ the Moore-Penrose pseudo-inverse of $A_d$, the unique solution $\bm{x} = A_d^+ \bm{v}_s$ always exists, since the rows of $A_d$ are linearly independent by definition \cite{meyer2023linearalgebra}.

If $\bm{x}_i \neq 0$ for all $\bm{x}_i \in \bm{x}$ then $\bm{v}_s$ is linear combination of the vectors in $B_d$ meaning that $\bm{v}_s \in V_d$. 
Since $B_d$ is directly defined after the roles of $d$ (Definition \ref{def:fd}), then the roles each role in $d$ must classify an entity in $s$. Based on this, we define how to deduce the descriptions satisfied by $s$ according to Definition \ref{def:dns-satisfies-definition}.

\begin{theorem} \label{theo:sandra-theorem}
    Given $s \in \mathcal{S}$, $d \in \mathcal{D}$, and $\bm{v}_s = \bm{f}_s(s)$ we have that 
    \begin{equation*}
        d \models s \Leftrightarrow \bm{v}_s \in V_d
    \end{equation*}
\end{theorem}

\begin{proof}
    With $d \models s$, then $\forall r \in d \ \exists e \in s . \psi(e) = r$ (i.e. exists an $e \in s$ classified by $r$ for all $r \in d$). If $e \in \mathcal{S}$, then $r \in \mathcal{D}$ and $r \models e$. Given Definition \ref{def:fs} ($\bm{f}_s$), $\bm{v}_s = \bm{f}_s(s)$ is obtained by aggregating the encoding of the roles that classify the individuals of $s$. Hence $\bm{v}_s \in V_d$, since it is linear combination of $B_d$ by construction. Thus, $d \models s \Rightarrow \bm{v}_s \in V_d$. To prove that $\bm{v}_s \in V_d \Rightarrow d \models s$, assume that $\bm{v}_s \in V_d$. Hence, $\bm{v}_s$ is linear combination of $B_d$. Given Definition \ref{def:fs}, $\forall r \in d \ \exists e \in s . \  \psi(e) \in r \lor r \models e$. It follows from Definition \ref{def:dns-satisfies-definition} (description satisfaction) that $d \models s$. Thus, $d \models s \Leftrightarrow \bm{f}_s(s) \in V_d$.
\end{proof}

\paragraph{$\mathds{P}$-\emph{sandra} (Probabilistic \emph{sandra})}
According to Theorem \ref{theo:sandra-theorem} each element of the vector $\bm{x}$ (with $\bm{x} = A_d^+ \bm{v}_s$) indicates the presence (or absence) of an entity $e \in s$ with $e \in r$ and $r \in d$. By applying the Heaviside function $\mathds{1}_{[>0]}$ to $\bm{x}$ we can interpret it as a boolean vector.

We model the satisfaction of a description as a multinomial probability distribution, hence the probability that $s$ satisfies $d$ is
\begin{equation} \label{eq:satisfy-prob}
    p(d \models s) = \frac{\sum\limits_{i = 0}^{|d|} \mathds{1}_{[>0]}(\bm{x})_i}{|d|}
\end{equation}

\begin{definition} \label{def:sandra-probability}
    Given $d \in \mathcal{D}$, $s \in \mathcal{S}$ we say that $s$ \textit{nearly}-satisfies $d$ with probability $\hat{p}$, written as $d \widetilde{\models}_{\hat{p}} s$, if $p(d \models s) = \hat{p}$.
\end{definition}

Given Definition \ref{def:sandra-probability}, we can now state Theorem \ref{theo:psandra-theorem}, which generalizes Theorem \ref{theo:sandra-theorem} to \textit{nearly}-satisfied descriptions according to Definition \ref{def:dns-nearly-satisfies-definition}.

\begin{theorem} \label{theo:psandra-theorem}
With $s \in \mathcal{S}$, $d \in \mathcal{D}$ then \begin{enumerate*}[label=(\roman*)]
  \item $d \widetilde{\models}_{\hat{p}} s \Rightarrow d \widetilde{\models} s$ with $\hat{p} = p(d \models s) > 0$;
  \item $d \widetilde{\models}_1 s \Rightarrow d \models s$;
  \item $d \widetilde{\models}_0 s \Rightarrow d \not\models s$.
\end{enumerate*}
\end{theorem}

\begin{proof}
\begin{enumerate}[label=(\roman*)]
    \item $d \widetilde{\models}_{\hat{p}} s$ then, according to Definition \ref{def:sandra-probability}, there is some $r \in d$ such that, given the coefficient $i$ that corresponds to $r$ in $\bm{x}$, $(\bm{x})_i > 0$. Hence $\bm{v}_s$ is a linear combination of $\bm{f}_d(r)$. Similarly to proof of Theorem \ref{theo:sandra-theorem}, given Definition \ref{def:fs} it can be proven that exists $e \in s$ with $e \in r$. Thus, from Definition \ref{def:dns-nearly-satisfies-definition}, $d \widetilde{\models} s$.
    \vspace{-0.2cm}
    \item $d \widetilde{\models}_1 s$ then $\sum_{i = 0}^{|d|} \mathds{1}_{[>0]}(\bm{x})_i = 1$ and $\bm{v}_s$ is linear combination of the vector representation of all the roles $r \in d$. Thus, $d \models s$ follows from Theorem \ref{theo:sandra-theorem}.
    \vspace{-0.2cm}
    \item $d \widetilde{\models}_0 s$ then $\sum_{i = 0}^{|d|} \mathds{1}_{[>0]}(\bm{x})_i = 0$ and $\bm{v}_s$ can not be expressed as a linear combination of any of the vector representation of the roles $r \in d$. Thus, $d \not\models s$ follows from Theorem \ref{theo:sandra-theorem}.
\end{enumerate}

\vspace{-0.4cm}
\end{proof}

Theorem \ref{theo:psandra-theorem} enables a crucial property of our formalization: to infer to which degree a description is satisfied by a situation, while retaining correctness with respect to Definitions \ref{def:dns-satisfies-definition} (satisfaction) and \ref{def:dns-nearly-satisfies-definition} (\textit{near}-satisfaction).

\paragraph{$d\mathds{P}$-\emph{sandra} (Differentiable $\mathds{P}$-\emph{sandra})}
By virtue of Theorem \ref{theo:psandra-theorem}, the process of deducing which descriptions are satisfied (or \textit{nearly}-satisfied) by a situation is a function $H: V \rightarrow \mathbb{R}^{|\mathcal{D}|}$, with
\begin{equation} \label{eq:deducing-function}
    H(s) = [ p(d_1 \models s) \Vert \cdots \Vert p(d_n \models s) ]
\end{equation}
with $d_1, \cdots, d_n \in \mathcal{D}$.

$H$ is differentiable with respect to the input vector $\bm{v}_s$ (it is composition of differentiable functions) and in particular $\nabla H = [ \delta(\bm{v}_s) \Vert \cdots \Vert \delta(\bm{v}_s)) ]$ where $\delta$ is Dirac's delta and $d_1, \cdots, d_n \in \mathcal{D}$.
Due to the use of the Heaviside function, the gradient is zero everywhere except for the descriptions with probability zero, since $\delta(x) = 0 \  \forall x \neq 0$.
This prevents an effective use of $H$ in Machine Learning methods that rely on gradients to learn the model's parameters, such as in Deep Learning, where the back-propagation algorithm is used \cite{lecun2015deeplearning}. 

To overcome this issue, we replace the Heaviside function $\mathds{1}_{[>0]}(\bm{x})$ with the $ReLU$ function \cite{glorot2011relu}. We argue, without any loss in generality, that the behaviour of the $ReLU$ function is comparable to the Heaviside function in our setting, since $ReLU(x) > 0 \Leftrightarrow x > 0$. Indeed, we can interpret the $ReLU$ function as a smooth version of $\mathds{1}_{>0}(\bm{x})$. Other functions could be used for the same purpose, such as continuous approximations of the Heaviside function, $tanh$, logistic function etc. In this case, we rely on $ReLU$ since empirically it suffers less from the exploding/vanishing gradient problem. Nonetheless, we will perform a systematic assessment of other functions in the future, as different functions might lead to different useful properties.

\subsection{Neuro-symbolic integration} \label{sec:nesy}
Since $d\mathds{P}$-sandra is differentiable with respect to its input vector, it can be used as a standard neural network layer. We define the function $\Tilde{H}$ by replacing the Heaviside function in $H$ (as defined in Equation \ref{eq:deducing-function}) with the $ReLU$ function.
The function $\Tilde{H}$ allows a seamless integration of a DnS-based ontology $\mathcal{O}$ within any arbitrary neural network $NN$. 

The $NN$ approximates the function $\bm{f}_s$ from a non-structured source. Consider $\bm{x}$ the output of a neural network, for instance the features extracted by a Convolutional Neural Network (CNN) from an image or the embedding of a sentence computed by a Transformer, then the $NN$ approximates the result of $\bm{f}_s(\bm{x})$ as if $\bm{x}$ was created from a structured source.

Hence, we can check if the situation represented by $\bm{x}$ satisfies $d$ by relying on Definition \ref{eq:satisfy-prob}. Moreover, since $\Tilde{H}$ is fully differentiable with respect to its input, it is possible to define an objective function that optimizes the output of the $NN$ such that the representation learned for $\bm{x}$ lies in $V_d$ as if it was produced by $\bm{f}_s$.

We highlight that our experiments (c.f. Section \ref{sec:experiments}) show that the representation learned by $NN$ (a vector $\bm{x}$ originating from any source) approximates the vector representation of a situation $s$ manually curated for modeling or representing that source, such as in a Knowledge Graph.
This provides us with a means to interpret the intermediate output of the $NN$, and better understand its internal representation, without influencing the process to solve the down-stream task. 
For example, given the image of a lung X-ray, \emph{sandra} infers all the possible diagnoses without compromising the freedom of the model to formulate the final prediction.

\section{Experiments}
\label{sec:experiments}
\begin{table*}[ht]
    \centering
    \hfill
    \subfloat[Results on the accuracy in the I-RAVEN dataset. Number of parameters of the baseline (first column) is compared with the addition of \textit{sandra}. The remaining columns refer to the specific configuration of the I-RAVEN dataset (number of shapes and their position).]{
        \begin{tabular}{ccccccccc}
        \toprule
         & \#Params & C & 2$\times$2 & 3$\times$3 & O-IC & O-IG & L-R & U-D  \\ \midrule
         Baseline & 205k & 26.85 & 14.55 & 12.15 & 12.3 & 13.4 & 11.85 & 13.15 \\ 
         \textit{sandra} & 275k & \textbf{45.75} & \textbf{16.15} & \textbf{14.1} & \textbf{14.8} & \textbf{14.85} & \textbf{13.05} & 13.15 \\
         \bottomrule
        \end{tabular}
        \label{tab:raven-results}
    }
    \hfill
    \subfloat[Results on the accuracy in R-FMNIST benchmark. The parameters in the CNN are $30k$ and $85.6k$ using \textit{sandra} while $268.7k$ and $278k$ using \textit{sandra} in the MLP.]{
        \small
        \begin{tabular}{lccc}
        \toprule
        Model & A & B & C \\ \midrule
        CNN & \textbf{16.02} & 17.21 & 43.13 \\
        sandra & 15.74 & \textbf{18.62} & \textbf{52.49} \\ \hline
        MLP & 14.56 & 16.64 & 31.34 \\
        sandra & \textbf{16.96} & \textbf{18.11} & \textbf{32.88} \\
        \bottomrule
        \end{tabular}
        \label{tab:domain-gen-results}
    }
    \hfill
    \label{tab:results}
    \vspace*{-2em}
    \caption{Experimental results.}
\end{table*}

We experiment the integration of $d\mathds{P}$-\textit{sandra} in different neural network architectures. Our goal is to assess whether \textit{sandra} allows to perform perspective-based reasoning without any performance loss or increase in computational complexity, as hyphotesized in Section \ref{sec:intro}.
We test our method on two benchmarks: I-RAVEN \cite{hu2021stratified}, a visual reasoning task, and RotatedFashionMNIST, a domain generalisation task based on FashionMNIST \cite{xiao2017fashionmnist}. The choice of the tasks is motivated by the suitability of the data to be modeled according to DnS with a reasonable effort.
All the experiments have been trained with the AdamW optimizer on a NVIDIA RTX3090 with 24Gb of RAM.

\subsection{Visual Reasoning} \label{sec:experiment:IRAVEN}
The I-RAVEN dataset, based on Raven Progressive Matrices (RPMs) is a repository of images for visual reasoning tasks, designed to examine abstract reasoning capabilities in neural networks \cite{hu2021stratified}. An RPM is a $3 \times 3$ grid of images where the last image has been removed. The task is to predict the removed image among $8$ possible alternatives \cite{malkinski2022deep}.
Each image might be composed of multiple (nested) figures, which are simple geometric shapes where the color, size and rotation are varied \cite{hu2021stratified}.
Recent approaches frame the task as a classification problem and tailor the model to the RPM problem. This has been done by structuring the models such that the relationship between the images is captured \cite{santoro2017simple,barrett2018measuring,zhang2019learning,he2023covitraven} or by reasoning upon the extracted visual features through neuro-symbolic methods \cite{xu2023abstract,hersche2023nesyraven}. Our approach extends the baseline model proposed by \cite{zhang2019learning} with \emph{sandra}.
The baseline model is composed by an LSTM that combines the visual features extracted by a CNN on each image. Details on the architecture are in the Appendix \footnote{Included in the additional material.}.
We integrate \emph{sandra} within the baseline model by adding a linear layer $W$ that projects the representation obtained by the CNN in the vector space $V$. $W$ is meant to approximate the function $\bm{f}_s$.

Given an image within an RPM, $s_R$ is the situation extracted from the XML provided by the dataset accordingly to Definition \ref{def:dns-description} (c.f. Appendix for details on the extraction process) and $\hat{\bm{x}}$ the vector representing the visual features of the image extracted by the CNN. We approximate $\bm{f}_s(s_R)$ by minimizing the binary cross entropy 
\begin{equation}
    \mathcal{L}_{BCE}(\hat{H}(\bm{f}_s(s)), \hat{H}(ReLU(W \hat{\bm{x}}))   
\end{equation}
where $\Tilde{H}$ is computed as explained in Section \ref{sec:nesy}.

Our model classifies an image using linear regression on $\hat{\bm{x}} \Vert \hat{H}(ReLU(W \hat{\bm{x}}))$. The loss function is hence
\begin{equation}
    \mathcal{L} = \mathcal{L}_{CE} + \mathcal{L}_{BCE}(\hat{H}(ReLU(W \hat{\bm{x}})), H(\bm{f}_s(s)))   
\end{equation}
where $\mathcal{L}_{CE}$ is the cross-entropy loss as defined in \cite{zhang2019learning}.

Our approach radically differs from the others since, by relying on \emph{sandra}, the visual features extracted from an image are geometrically constrained such that it is possible to infer valid descriptions from them. The model learns to accurately classify an image, while capturing the semantics of the ontology in its internal representation. The influence of \emph{sandra} is a regularization with respect to a DnS-based ontology. We manually compile a DnS-based ontology (detailed in Appendix) that models the RAVEN dataset. Each image is then converted into a situation compliant to the ontology.
Relying on \emph{sandra} allows the inference of which shapes are contained within a figure. 
A comparison of \textit{sandra} with the baseline is shown in Table \ref{tab:raven-results}, we report the of number of parameters of the network (first column) and accuracy (the other columns) of the classification. Each model has been trained using a batch size of $32$, a learning rate of $0.001$ and the gradient is clipped to have a norm $\leq 0.5$. 
The results show that integrating \emph{sandra} does not compromise performances (improving them in some cases) nor it requires a significant increase in the number of parameters.
We also compare \textit{sandra} with other approaches tailored for this task and for the sake of space, the comparison is reported in the Appendix. We remark that, despite its generality, our method is competitive with other methods \footnote{We remind that our primary goal is to test our hypothesis rather than solving the downstream task.}.

\subsection{Domain Generalization} \label{sec:experiment:fashionMNIST}
\begin{figure}[htb]
    \centering
    \setlength{\belowcaptionskip}{-10pt}
    \includegraphics[scale=0.8]{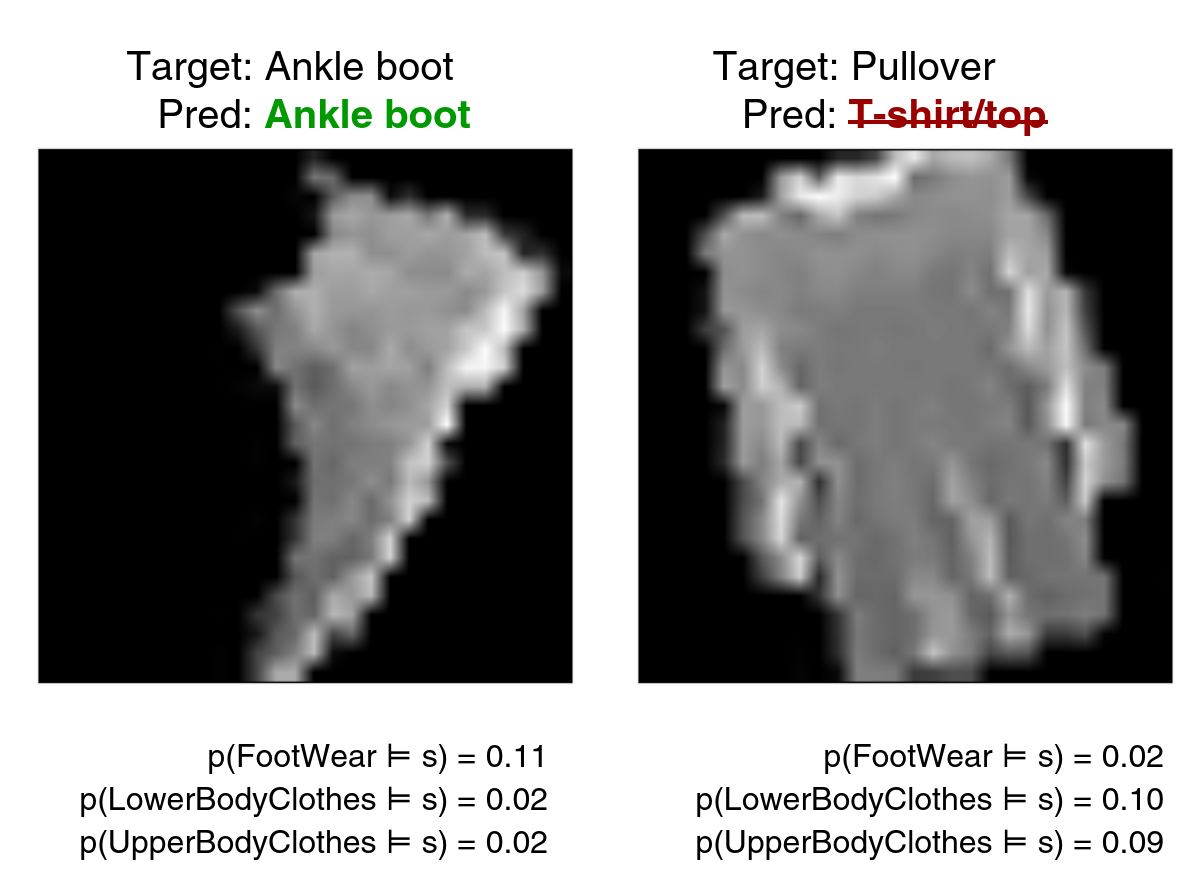}
    \vspace*{-0.5em}
    \caption{Example of two images and the descriptions they satisfy. The first image has a high probability of satisfying the \textit{FootWear} description, which is reasonable considering the target label. In the second image, we can understand that the wrong classification is associated with the similar probability given to the \textit{LowerBodyClothes} and \textit{UpperBodyClothes} descriptions, leading to an unreliable result.}
    \label{fig:interpretation}
\end{figure}

The domain generalization task aims at testing how a model is able to learn representations that are independent of a specific domain, i.e. they are effective on distributions never seen during the training process \cite{zhou2022domain}.
This includes, for example, correctly classifying images seen from a different angle than the one in the training set. Many approaches have been investigated, from learning representations that specifically tackles the problem \cite{mahajan2021domain}, to different training techniques \cite{vedatam2021domaingen}. A wide selection of architectures and techniques is analysed in \cite{zhou2023domaingenurvey}.
Similarly to \cite{mahajan2021domain}, we rely on the Rotated-FashionMNIST (R-FMNIST) dataset. R-FMNIST is an image classification task where the training images are rotated in three configurations: $A = \{30^\circ, 45^\circ\}$, $B \equiv A \cup \{ 60^\circ \}$, $C \equiv B \cup \{ 15^\circ, 75^\circ \}$. The testing set images are rotated by $\{0^\circ, 90^\circ\}$.

The final prediction is computed using linear regression over the features extracted by a CNN or a MLP. The architectures are described in the Appendix. We manually create a DnS-based ontology, detailed in the Appendix, that models the dataset. 
Table \ref{tab:domain-gen-results} describes the results in terms of number of parameters of the network (in the caption) and accuracy of the classification. Each model has been trained with a batch size of $32$ and a fixed learning rate of $1e^{-4}$. The use of \emph{sandra} proves to be beneficial also in this experiment even though the images of R-FMNIST are only described with a single label, leading to a fairly simple ontology.

These experiments allow us to empirically show the \textit{sandra} improves the interpretability of the model and of its results, as addressed in Section \ref{sec:method}. Figure \ref{fig:interpretation} shows the effect of \textit{sandra} in the interpretation of two images, one correctly labeled and one misclassified. By looking at the inferred descriptions, it is possible to interpret the motivation of the mistake.



\section{Discussion and Future Works}
\label{sec:discussion}
\emph{Sandra} approaches the problem of reasoning on different persepectives by formalizing DnS in a vector space.
Through the formulation described in Section \ref{sec:method} it is possible to infer the descriptions satisfied by a situation. Most importantly, it is possible to quantify the degree of satisfiability as a probability distribution, enabling the inference of valid interpretations of a situation regardless of their compatibility and of the amount of information available. This is an important aspect that finds application in many domains where rigorous reasoning with limited information is a key requirement such as robotics, medicine or jurisprudence.
We discuss some limitations and interesting aspects that are worth further investigation.
\paragraph{DnS-shape Compatibility}
\emph{Sandra}'s dependency on DnS ontologies, which underpins $V$, presents a non-trivial limitation. This is relevant if one wants to integrate it with ontologies using different (if any) ODP.
However, we argue that it is possible to find an isomorphism between any arbitrary ontology and a DnS ontology. Informally, we can use reification. Each binary predicate (property) $p$ can become a class expressing a n-ary relation, whose arguments are its domain and range. Any axiom involving the property $p$ can be reified as a description including $p$ as role. 
When reversed - as expected - this transformation may be backward inconsistent, since the inferred satisfied descriptions may correspond to disjoint classes in the original ontology. 
We will explore how to formally define such isomorphic transformation, alongside its properties and limitations. Another consideration is that the existence of an ontology compatible with DnS does not always guarantee improved performance. We need further research to identify the most effective DnS transformation patterns for integration with \emph{sandra}. We plan to study \emph{sandra}'s behavior with other ontologies, particularly with complex ones, to find correlations between ontology design solutions and \emph{sandra}'s performance.
\paragraph{Ontology Complexity and Scalability}
Another issue is the inherent computational complexity. The function $\bm{f}_s$ is linear in space with the dimension of the ontology. While this might seem an ideal condition, as the ontology gets more complex (e.g. thousands of descriptions) this might result in a large overhead, especially if $\bm{f}_s$ is approximated through a neural network. Further research is required to identify if it is possible to obtain more compact representations, for instance by exploiting the hierarchical structure of the ontology or by exploring methods that pre-process the ontology removing non-essential elements. 
\paragraph{Methodological Improvements}
The definitions of $\bm{f}_d$ and $\bm{f}_s$ might be improved to obtain additional useful properties. For instance, our current method overlooks a procedure to retrieve the original situation $s$ from $\bm{f}_s$. This precludes useful outcomes, such as uniquely identifying which entity of $s$ has been classified by which role in $d$. An interesting approach is to extend the method of Section \ref{sec:method} to some parts of Description Logic as well. In that case, it might be possible to benefit from additional features that have been formalized in the ontology but that are unexploited since they are missing in the DnS pattern.
\paragraph{Neurosymbolic integration}
The results of Section \ref{sec:experiments} provide a positive insight on how \emph{sandra} can be beneficial to neural architectures. Nonetheless, the variance in performance gains between different tasks requires a systematic analysis, in order to better understand which kind of architectures can benefit the most from \emph{sandra} and how they should be integrated. Investigating the ODPs that lead to increased performances is an important step towards tighter integration between Machine Learning (ML) and  Knowledge Representation (KR). Notably, this aspect has the potential to catalyse a novel paradigm in the realm of KR, since different ML techniques might behave differently with different ontologies. 

\section{Related Works}
\label{sec:related}


The idea of using geometric representation for cognitive theory was originally proposed by Peter Gärdenfors \shortcite{gardenfors2004conceptual} as an intermediate representation layer between sub-symbolic and symbolic knowledge. A conceptual space depicts concepts as convex regions. Each region is shaped by the attributes that characterise that concept. Our method follows the  intuition of conceptual spaces.
Two main problems on conceptual spaces are highlighted by \cite{bouraoui2022integrating}. One is the derivation of region-based representations for concepts that is difficult when data comes from real-world scenarios; another issue is that the representations in conceptual spaces are not amenable to expressing relational knowledge. Given the integration of a neural network as a situation approximator, as proposed in Section \ref{sec:nesy}, and the DnS formalization ( Section \ref{sec:dns}), our method overcomes both limitations. Other approaches that implement conceptual spaces, such as \cite{bouraoui2022integrating}, propose to interpret embeddings as conceptual spaces. Others approximate ontologies based on their graph representation or rely on specific neural architectures, like Logic Tensor Networks to obtain deductive inferences \cite{chen2021owl2vec,ebrahimi2021deepdeductivereasoner}. Differently to these approaches, our representation is not an approximation of the ontology. The ontology is used to obtain a vector space in which concepts are represented. This is similar to the approach described in \cite{ozccep2020cone}, where geometrical properties are exploited to obtain sound knowledge representation methods.

Other related approaches integrate symbolic representations in geometrical spaces to support reasoning such as in \cite{geh2023dpasp} where the authors complement Answer Set Programming with neural networks; in \cite{manhaeve2018deepproblog} ProbLog is extended to the use of neural predicates; and \cite{van2022nesi} implements approximate logical inference. These approaches are inherently dependent on probabilistic logic, suffering from a lack of inconsistency tolerance. 

\section{Conclusion}
\label{sec:conclusion}
We contribute a novel neuro-symbolic approach named \emph{sandra} to address perspective-based reasoning based on the DnS ontology design pattern. Our experiments show that \emph{sandra} brings benefits to the representation learning process without loss of performances or increase in computational complexity. 



\section*{Acknowledgments}
This project has received funding from the FAIR – Future Artificial Intelligence Research foundation as part of the grant agreement MUR n. 341 and from the European Union’s Horizon 2020 research and innovation programme under grant agreement No 101004746.

\bibliographystyle{named}
\bibliography{ijcai24}

\begin{thebibliography}{}

\bibitem[\protect\citeauthoryear{Asprino \bgroup \em et al.\egroup }{2022}]{asprino2022sparqlanything}
Luigi Asprino, Enrico Daga, Aldo Gangemi, and Paul Mulholland.
\newblock Knowledge graph construction with a fa\c{c}ade: A unified method to access heterogeneous data sources on the web.
\newblock {\em ACM Trans. Internet Technol.}, 2022.

\bibitem[\protect\citeauthoryear{Barrett \bgroup \em et al.\egroup }{2018}]{barrett2018measuring}
David Barrett, Felix Hill, Adam Santoro, Ari Morcos, and Timothy Lillicrap.
\newblock Measuring abstract reasoning in neural networks.
\newblock In {\em International conference on machine learning}, pages 511--520. PMLR, 2018.

\bibitem[\protect\citeauthoryear{Beckett \bgroup \em et al.\egroup }{2014}]{beckett2014ttl}
David Beckett, Tim Berners-Lee, Eric Prud’hommeaux, and Gavin Carothers.
\newblock Rdf 1.1 turtle.
\newblock {\em World Wide Web Consortium}, pages 18--31, 2014.

\bibitem[\protect\citeauthoryear{Bikakis \bgroup \em et al.\egroup }{2021}]{10.3233/SW-200416}
Antonis Bikakis, Beatrice Markhoff, Alessandro Mosca, Stephane Jean, Eero Hyv\"{o}nen, Francesco Beretta, Antonis Bikakis, Eero Hyvonen, St\'{e}phane Jean, Beatrice Markhoff, and Alessandro Mosca.
\newblock A challenge for historical research: Making data fair using a collaborative ontology management environment (ontome).
\newblock {\em Semant. Web}, 12(2):279–294, jan 2021.

\bibitem[\protect\citeauthoryear{Bouraoui \bgroup \em et al.\egroup }{2022}]{bouraoui2022integrating}
Zied Bouraoui, Victor Gutierrez~Basulto, and Steven Schockaert.
\newblock Integrating ontologies and vector space embeddings using conceptual spaces.
\newblock {\em International Research School in Artificial Intelligence in Bergen}, 99:3, 2022.

\bibitem[\protect\citeauthoryear{{Boyan Brodaric and Femke Reitsma and Yi Qiang}}{2008}]{723935}
{Boyan Brodaric and Femke Reitsma and Yi Qiang}.
\newblock {SKIing with DOLCE : toward an e-Science knowledge infrastructure}.
\newblock In {Carola Eschenbach and Michael Gruninger}, editor, {\em {Frontiers in Artificial Intelligence and Applications}}, volume {183}, pages {208--219}. {IOS Press}, {2008}.

\bibitem[\protect\citeauthoryear{Chen \bgroup \em et al.\egroup }{2021}]{chen2021owl2vec}
Jiaoyan Chen, Pan Hu, Ernesto Jim{\'{e}}nez{-}Ruiz, Ole~Magnus Holter, Denvar Antonyrajah, and Ian Horrocks.
\newblock Owl2vec*: embedding of {OWL} ontologies.
\newblock {\em Mach. Learn.}, 110(7):1813--1845, 2021.

\bibitem[\protect\citeauthoryear{Ebrahimi \bgroup \em et al.\egroup }{2021}]{ebrahimi2021deepdeductivereasoner}
Monireh Ebrahimi, Aaron Eberhart, Federico Bianchi, and Pascal Hitzler.
\newblock Towards bridging the neuro-symbolic gap: deep deductive reasoners.
\newblock {\em Appl. Intell.}, 51(9):6326--6348, 2021.

\bibitem[\protect\citeauthoryear{Fillmore and others}{2006}]{fillmore2006frame}
Charles~J Fillmore et~al.
\newblock Frame semantics.
\newblock {\em Cognitive linguistics: Basic readings}, 34:373--400, 2006.

\bibitem[\protect\citeauthoryear{Gangemi and Mika}{2003}]{gangemi2003understanding}
Aldo Gangemi and Peter Mika.
\newblock Understanding the semantic web through descriptions and situations.
\newblock In {\em OTM Confederated International Conferences" On the Move to Meaningful Internet Systems"}, pages 689--706. Springer, 2003.

\bibitem[\protect\citeauthoryear{Gardenfors}{2004}]{gardenfors2004conceptual}
Peter Gardenfors.
\newblock {\em Conceptual spaces: The geometry of thought}.
\newblock MIT press, 2004.

\bibitem[\protect\citeauthoryear{Geh \bgroup \em et al.\egroup }{2023}]{geh2023dpasp}
Renato~Lui Geh, Jonas Gon{\c{c}}alves, Igor~Cataneo Silveira, Denis~Deratani Mau{\'a}, and Fabio~Gagliardi Cozman.
\newblock dpasp: A comprehensive differentiable probabilistic answer set programming environment for neurosymbolic learning and reasoning.
\newblock {\em arXiv preprint arXiv:2308.02944}, 2023.

\bibitem[\protect\citeauthoryear{Glimm \bgroup \em et al.\egroup }{2014}]{glimm2014hermit}
Birte Glimm, Ian Horrocks, Boris Motik, Giorgos Stoilos, and Zhe Wang.
\newblock Hermit: An {OWL} 2 reasoner.
\newblock {\em J. Autom. Reason.}, 53(3):245--269, 2014.

\bibitem[\protect\citeauthoryear{Glorot \bgroup \em et al.\egroup }{2011}]{glorot2011relu}
Xavier Glorot, Antoine Bordes, and Yoshua Bengio.
\newblock Deep sparse rectifier neural networks.
\newblock In Geoffrey~J. Gordon, David~B. Dunson, and Miroslav Dud{\'{\i}}k, editors, {\em Proceedings of the Fourteenth International Conference on Artificial Intelligence and Statistics, {AISTATS} 2011, Fort Lauderdale, USA, April 11-13, 2011}, volume~15 of {\em {JMLR} Proceedings}, pages 315--323. JMLR.org, 2011.

\bibitem[\protect\citeauthoryear{He \bgroup \em et al.\egroup }{2023}]{he2023covitraven}
Wentao He, Jialu Zhang, Jianfeng Ren, Ruibin Bai, and Xudong Jiang.
\newblock Hierarchical convit with attention-based relational reasoner for visual analogical reasoning.
\newblock In Brian Williams, Yiling Chen, and Jennifer Neville, editors, {\em Thirty-Seventh {AAAI} Conference on Artificial Intelligence, {AAAI} 2023, Thirty-Fifth Conference on Innovative Applications of Artificial Intelligence, {IAAI} 2023, Thirteenth Symposium on Educational Advances in Artificial Intelligence, {EAAI} 2023, Washington, DC, USA, February 7-14, 2023}, pages 22--30. {AAAI} Press, 2023.

\bibitem[\protect\citeauthoryear{Hersche \bgroup \em et al.\egroup }{2023}]{hersche2023nesyraven}
Michael Hersche, Mustafa Zeqiri, Luca Benini, Abu Sebastian, and Abbas Rahimi.
\newblock A neuro-vector-symbolic architecture for solving raven's progressive matrices.
\newblock {\em Nat. Mac. Intell.}, 5(4):363--375, 2023.

\bibitem[\protect\citeauthoryear{Horridge and Patel-Schneider}{2009}]{horridge2009manchester}
Matthew Horridge and Peter~F Patel-Schneider.
\newblock Owl 2 web ontology language manchester syntax.
\newblock {\em W3C Working Group Note}, 2009.

\bibitem[\protect\citeauthoryear{Hu \bgroup \em et al.\egroup }{2021}]{hu2021stratified}
Sheng Hu, Yuqing Ma, Xianglong Liu, Yanlu Wei, and Shihao Bai.
\newblock Stratified rule-aware network for abstract visual reasoning.
\newblock In {\em Proceedings of the AAAI Conference on Artificial Intelligence}, volume~35, pages 1567--1574, 2021.

\bibitem[\protect\citeauthoryear{Kamide}{2022}]{kamide2022fuzzydl}
Norihiro Kamide.
\newblock Reasoning with inconsistency-tolerant fuzzy description logics.
\newblock In Ana~Paula Rocha, Luc Steels, and H.~Jaap van~den Herik, editors, {\em Proceedings of the 14th International Conference on Agents and Artificial Intelligence, {ICAART} 2022, Volume 3, Online Streaming, February 3-5, 2022}, pages 63--74. {SCITEPRESS}, 2022.

\bibitem[\protect\citeauthoryear{LeCun \bgroup \em et al.\egroup }{2015}]{lecun2015deeplearning}
Yann LeCun, Yoshua Bengio, and Geoffrey~E. Hinton.
\newblock Deep learning.
\newblock {\em Nat.}, 521(7553):436--444, 2015.

\bibitem[\protect\citeauthoryear{Mahajan \bgroup \em et al.\egroup }{2021}]{mahajan2021domain}
Divyat Mahajan, Shruti Tople, and Amit Sharma.
\newblock Domain generalization using causal matching.
\newblock In {\em International Conference on Machine Learning}, pages 7313--7324. PMLR, 2021.

\bibitem[\protect\citeauthoryear{Ma{\l}ki{\'n}ski and Ma{\'n}dziuk}{2022}]{malkinski2022deep}
Miko{\l}aj Ma{\l}ki{\'n}ski and Jacek Ma{\'n}dziuk.
\newblock Deep learning methods for abstract visual reasoning: A survey on raven's progressive matrices.
\newblock {\em arXiv preprint arXiv:2201.12382}, 2022.

\bibitem[\protect\citeauthoryear{Manhaeve \bgroup \em et al.\egroup }{2018}]{manhaeve2018deepproblog}
Robin Manhaeve, Sebastijan Dumancic, Angelika Kimmig, Thomas Demeester, and Luc De~Raedt.
\newblock Deepproblog: Neural probabilistic logic programming.
\newblock {\em Advances in neural information processing systems}, 31, 2018.

\bibitem[\protect\citeauthoryear{Meyer and Stewart}{2023}]{meyer2023linearalgebra}
Carl~D Meyer and Ian Stewart.
\newblock {\em Matrix analysis and applied linear algebra}.
\newblock SIAM, 2023.

\bibitem[\protect\citeauthoryear{Minsky}{1974}]{minsky1974framework}
Marvin Minsky.
\newblock A framework for representing knowledge, 1974.

\bibitem[\protect\citeauthoryear{{\"O}z{\c{c}}ep \bgroup \em et al.\egroup }{2020}]{ozccep2020cone}
{\"O}zg{\"u}r~L{\"u}tf{\"u} {\"O}z{\c{c}}ep, Mena Leemhuis, and Diedrich Wolter.
\newblock Cone semantics for logics with negation.
\newblock In {\em IJCAI}, pages 1820--1826, 2020.

\bibitem[\protect\citeauthoryear{Porzel and Cangalovic}{2020}]{Porzel2020WhatSY}
Robert Porzel and Vanja~Sophie Cangalovic.
\newblock What say you: An ontological representation of imperative meaning for human-robot interaction.
\newblock In {\em Joint Ontology Workshops}, 2020.

\bibitem[\protect\citeauthoryear{Santoro \bgroup \em et al.\egroup }{2017}]{santoro2017simple}
Adam Santoro, David Raposo, David~G Barrett, Mateusz Malinowski, Razvan Pascanu, Peter Battaglia, and Timothy Lillicrap.
\newblock A simple neural network module for relational reasoning.
\newblock {\em Advances in neural information processing systems}, 30, 2017.

\bibitem[\protect\citeauthoryear{Santoro \bgroup \em et al.\egroup }{2018}]{santoro2018raven}
Adam Santoro, Felix Hill, David G.~T. Barrett, Ari~S. Morcos, and Timothy~P. Lillicrap.
\newblock Measuring abstract reasoning in neural networks.
\newblock In Jennifer~G. Dy and Andreas Krause, editors, {\em Proceedings of the 35th International Conference on Machine Learning, {ICML} 2018, Stockholmsm{\"{a}}ssan, Stockholm, Sweden, July 10-15, 2018}, volume~80 of {\em Proceedings of Machine Learning Research}, pages 4477--4486. {PMLR}, 2018.

\bibitem[\protect\citeauthoryear{Scherp \bgroup \em et al.\egroup }{2009}]{10.1145/1597735.1597760}
Ansgar Scherp, Thomas Franz, Carsten Saathoff, and Steffen Staab.
\newblock F--a model of events based on the foundational ontology dolce+dns ultralight.
\newblock In {\em Proceedings of the Fifth International Conference on Knowledge Capture}, K-CAP '09, page 137–144, New York, NY, USA, 2009. Association for Computing Machinery.

\bibitem[\protect\citeauthoryear{van Krieken \bgroup \em et al.\egroup }{2022}]{van2022nesi}
Emile van Krieken, Thiviyan Thanapalasingam, Jakub~M Tomczak, Frank van Harmelen, and Annette~ten Teije.
\newblock A-nesi: A scalable approximate method for probabilistic neurosymbolic inference.
\newblock {\em arXiv preprint arXiv:2212.12393}, 2022.

\bibitem[\protect\citeauthoryear{Vedantam \bgroup \em et al.\egroup }{2021}]{vedatam2021domaingen}
Ramakrishna Vedantam, David Lopez{-}Paz, and David~J. Schwab.
\newblock An empirical investigation of domain generalization with empirical risk minimizers.
\newblock In Marc'Aurelio Ranzato, Alina Beygelzimer, Yann~N. Dauphin, Percy Liang, and Jennifer~Wortman Vaughan, editors, {\em Advances in Neural Information Processing Systems 34: Annual Conference on Neural Information Processing Systems 2021, NeurIPS 2021, December 6-14, 2021, virtual}, pages 28131--28143, 2021.

\bibitem[\protect\citeauthoryear{Vossen and Fokkens}{2022}]{vossen2022perspective}
Piek Vossen and Antske Fokkens.
\newblock {\em Creating a More Transparent Internet}.
\newblock Studies in Natural Language Processing. Cambridge University Press, 2022.

\bibitem[\protect\citeauthoryear{Xiao \bgroup \em et al.\egroup }{2017}]{xiao2017fashionmnist}
Han Xiao, Kashif Rasul, and Roland Vollgraf.
\newblock Fashion-mnist: a novel image dataset for benchmarking machine learning algorithms.
\newblock {\em arXiv preprint arXiv:2308.02944}, 2017.

\bibitem[\protect\citeauthoryear{Xu \bgroup \em et al.\egroup }{2023}]{xu2023abstract}
Jingyi Xu, Tushar Vaidya, Yufei Wu, Saket Chandra, Zhangsheng Lai, and Kai Fong~Ernest Chong.
\newblock Abstract visual reasoning: An algebraic approach for solving raven's progressive matrices.
\newblock In {\em Proceedings of the IEEE/CVF Conference on Computer Vision and Pattern Recognition}, pages 6715--6724, 2023.

\bibitem[\protect\citeauthoryear{Zhang \bgroup \em et al.\egroup }{2019a}]{zhang2019learning}
Chi Zhang, Feng Gao, Baoxiong Jia, Yixin Zhu, and Song{-}Chun Zhu.
\newblock {RAVEN:} {A} dataset for relational and analogical visual reasoning.
\newblock In {\em {IEEE} Conference on Computer Vision and Pattern Recognition, {CVPR} 2019, Long Beach, CA, USA, June 16-20, 2019}, pages 5317--5327. Computer Vision Foundation / {IEEE}, 2019.

\bibitem[\protect\citeauthoryear{Zhang \bgroup \em et al.\egroup }{2019b}]{chizhang2019raven}
Chi Zhang, Baoxiong Jia, Feng Gao, Yixin Zhu, Hongjing Lu, and Song{-}Chun Zhu.
\newblock Learning perceptual inference by contrasting.
\newblock In Hanna~M. Wallach, Hugo Larochelle, Alina Beygelzimer, Florence d'Alch{\'{e}}{-}Buc, Emily~B. Fox, and Roman Garnett, editors, {\em Advances in Neural Information Processing Systems 32: Annual Conference on Neural Information Processing Systems 2019, NeurIPS 2019, December 8-14, 2019, Vancouver, BC, Canada}, pages 1073--1085, 2019.

\bibitem[\protect\citeauthoryear{Zhang \bgroup \em et al.\egroup }{2021}]{zhang2021raven}
Chi Zhang, Baoxiong Jia, Song{-}Chun Zhu, and Yixin Zhu.
\newblock Abstract spatial-temporal reasoning via probabilistic abduction and execution.
\newblock In {\em {IEEE} Conference on Computer Vision and Pattern Recognition, {CVPR} 2021, virtual, June 19-25, 2021}, pages 9736--9746. Computer Vision Foundation / {IEEE}, 2021.

\bibitem[\protect\citeauthoryear{Zheng \bgroup \em et al.\egroup }{2019}]{zheng2019raven}
Kecheng Zheng, Zheng{-}Jun Zha, and Wei Wei.
\newblock Abstract reasoning with distracting features.
\newblock In Hanna~M. Wallach, Hugo Larochelle, Alina Beygelzimer, Florence d'Alch{\'{e}}{-}Buc, Emily~B. Fox, and Roman Garnett, editors, {\em Advances in Neural Information Processing Systems 32: Annual Conference on Neural Information Processing Systems 2019, NeurIPS 2019, December 8-14, 2019, Vancouver, BC, Canada}, pages 5834--5845, 2019.

\bibitem[\protect\citeauthoryear{Zhou \bgroup \em et al.\egroup }{2022}]{zhou2022domain}
Kaiyang Zhou, Ziwei Liu, Yu~Qiao, Tao Xiang, and Chen~Change Loy.
\newblock Domain generalization: A survey.
\newblock {\em IEEE Transactions on Pattern Analysis and Machine Intelligence}, 2022.

\bibitem[\protect\citeauthoryear{Zhou \bgroup \em et al.\egroup }{2023}]{zhou2023domaingenurvey}
Kaiyang Zhou, Ziwei Liu, Yu~Qiao, Tao Xiang, and Chen~Change Loy.
\newblock Domain generalization: {A} survey.
\newblock {\em {IEEE} Trans. Pattern Anal. Mach. Intell.}, 45(4):4396--4415, 2023.

\bibitem[\protect\citeauthoryear{Zhuo and Kankanhalli}{2021}]{zhuo2021raven}
Tao Zhuo and Mohan~S. Kankanhalli.
\newblock Effective abstract reasoning with dual-contrast network.
\newblock In {\em 9th International Conference on Learning Representations, {ICLR} 2021, Virtual Event, Austria, May 3-7, 2021}. OpenReview.net, 2021.

\bibitem[\protect\citeauthoryear{Zhuo \bgroup \em et al.\egroup }{2021}]{taozhuo2021raven}
Tao Zhuo, Qiang Huang, and Mohan~S. Kankanhalli.
\newblock Unsupervised abstract reasoning for raven's problem matrices.
\newblock {\em {IEEE} Trans. Image Process.}, 30:8332--8341, 2021.

\end{thebibliography}

\clearpage
\appendix
\label{app}
\section*{Appendix}

The Appendix provides the details omitted from the manuscript due to space limitations and the links to the implementations to reproduce the results. 
In Section \ref{app:experiments} we explain in details the architectures and ontologies used in Section \ref{sec:experiments} (experiments). 
The python implementation of \textit{sandra} is available at \url{https://anonymous.4open.science/r/sandra-C3D3/}.


\section{Experiments} \label{app:experiments}
This section provides additional details on the experiments performed in Section \ref{sec:experiments}. Section \ref{app:raven} describes the model's architecture and the ontology used in Section \ref{sec:experiment:IRAVEN}. 
Section \ref{app:fashion-mnist} we describes the models' architectures and the ontology used in Section \ref{app:fashion-mnist}.

\subsection{Visual Reasoning} \label{app:raven}
The original I-RAVEN dataset includes three main types of entities: \textit{Matrices}, \textit{Panels}, and \textit{Figures}.
In I-RAVEN a \textit{Matrix} is a set of 9 Panels. A \textit{Panel} is a component of the \textit{Matrix}. Each \textit{Panel} has a numeric value (from 1 to 9), which encodes its position in the \textit{Matrix}, from top left to bottom right. A \textit{Panel} could include a number N of \textit{Figures}. Each Figure has a \textit{Shape}, a \textit{Rotation Angle}, a \textit{Color}, and a \textit{Size}.

\begin{figure}[ht]
    \centering
    \setlength{\belowcaptionskip}{-10pt}
    \includegraphics[width=.7\columnwidth]{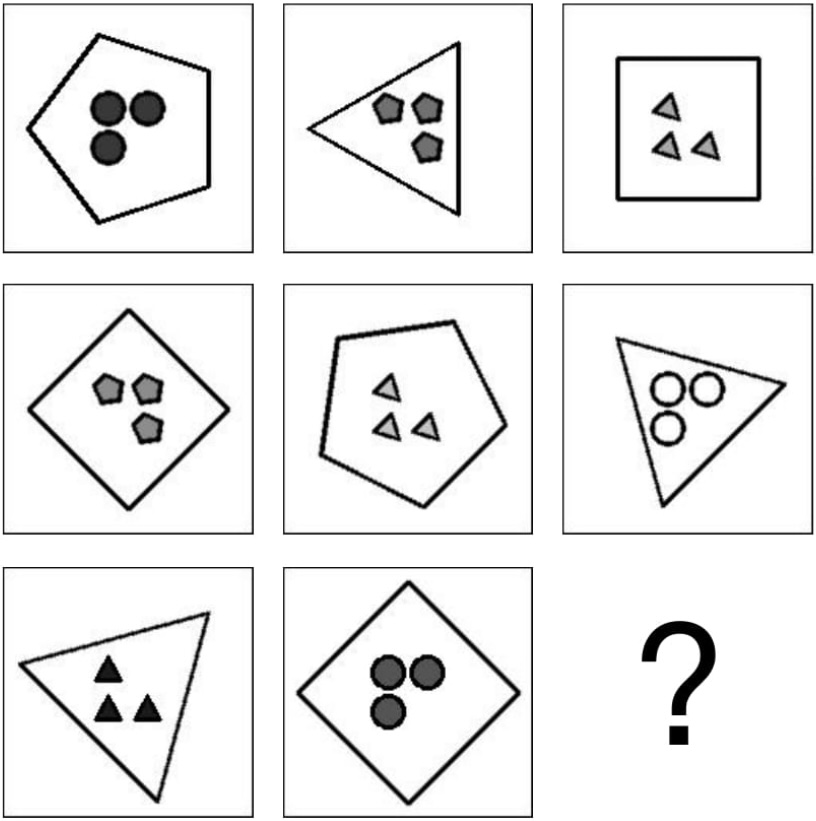}
    \caption{Raven Progressive Matrix example. The RPM is composed by $9$ numbered Panels ($3 \times 3$), the last one will be determined based on the $8$ explicit panels. Each \textit{Panel} can include a variable number of \textit{Figures} with some \textit{Shape}, \textit{Rotation Angle}, \textit{Color}, and \textit{Size}.}
    \label{fig:rpm}
\end{figure}

For example in Figure \ref{fig:rpm}, the \textit{Matrix} is composed of $9$ \textit{Panels}. The \textit{Shape} in the first \textit{Panel} is a \textit{Figure} containing a pentagon and composed of 3 small inner circles. The \textit{Color} of the inner circles is black. The outer pentagon is rotated.

\subsubsection{I-RAVEN ontology} \label{app:raven:ontology}
We reverse-engineer the I-RAVEN conceptual module to model the I-RAVEN intensional layer. We define the set of components that can appear in an image (see Figure \ref{fig:rpm}) based on the I-RAVEN official code~\footnote{Available at \url{https://github.com/husheng12345/SRAN/}}. We model a set of descriptions based on these components. 

Each RPM configurations is modeled as a situation. The situation is compiled starting from the original XML generated alongside the images, which contains structured information on the RPM such as the specific composition of each panel.

We use SPARQL Anything \cite{asprino2022sparqlanything} to interpret and transform the XML data into situations, serialized in RDF using Turtle notation \cite{beckett2014ttl}. SPARQL Anything is a querying tool for Semantic Web re-engineering. It allows users to query several data formats with the SPARQL query language. Through the use of the \texttt{CONSTRUCT} clause it is possible to compile a Knowledge Graph from the data formats supported by it.

The complexity of the ontology accommodates various meta-levels of deductions pertaining to different situations. For example, according to the above description, we add the axiom \texttt{rv:hasComponent \textbf{some} rv:Figure} to the \texttt{rv:Panel} class. This results in \texttt{Panel} being a description ($\texttt{rv:Panel} \in \mathcal{D}$) that includes the description \texttt{rv:Figure} ($\texttt{rv:Figure} \in \mathcal{D}$ and $\texttt{rv:Figure} \in \texttt{rv:Panel}$).
\texttt{:Figure} is modeled as a description which includes the roles \texttt{:Angle}, \texttt{:Color} and \texttt{:Size}. In order to model the semantics within the I-RAVEN dataset and to allow for better understanding of the data, we model the combinatorial permutation of figures within panels as descriptions as well.
For instance, a figure containing a circle satisfies the description $\texttt{rv:FT5} \in \mathcal{D}$, with $\texttt{rv:T5} \in \texttt{rv:FT5}$. The role $\texttt{rv:T5} \in \mathcal{R}$ is instantiated in a situation if the situation (i.e. the figure) contains a circle.

The I-RAVEN ontology models a description by declaring universal restrictions on the classes representing a role. This is done through the use of the \texttt{rdfs:subClassOf} predicate. For instance, \texttt{rv:FT5 rdfs:subClassOf \textbf{some} rv:T5}, which translates to $(\texttt{rv:FT5}) \sqsubseteq \forall(\texttt{rdfs:subClassOf}).(\texttt{rv:T5})$ in Description Logic. In total, the ontology contains $144$ classes ($|\mathcal{D} \cup \mathcal{R}| = 144$).
Table \ref{tab:iraven-ontology-classes} provides a detailed description of all the descriptions and roles in the ontology.

\subsubsection{Model Architecture}
The model's architecture used for the visual reasoning experiments is a straightforward CNN followed by a single-layer LSTM. The last hidden state of the LSTM is used to perform the classification using a linear classification head. The architecture is described in detail in Table \ref{tab:raven-architecture}.

\begin{table}[htb]
    \centering
    \begin{tabular}{l}
        \toprule
        3x3 conv., 16 filters, 2 stride, BN, ReLU \\
        3x3 conv., 16 filters, 2 stride, BN, ReLU \\
        3x3 conv., 16 filters, 2 stride, BN, ReLU \\
        3x3 conv., 16 filters, 2 stride, BN, ReLU \\
        \textit{Linear layer for projection in sandra's $V$} \\
        LSTM, 1 layer, $128$ hidden dimension \\
        Linear layer \\ \midrule
        Total of $205$ K parameters \\
        Total of $275$ K parameters with \emph{sandra} \\
        \bottomrule
    \end{tabular}
    \caption{Architecture used for the experiments in the I-RAVEN dataset.}
    \label{tab:raven-architecture}
\end{table}

\subsubsection{Comparison with state of the art models}
In Table \ref{tab:raven-results-app} a comparison of the results of Section \ref{sec:experiments} with recent works is reported. An estimate of the number of parameters is also described. \emph{Sandra} performs better than the baseline but is outperformed by the other approaches. We remark that the other approaches are optimized to solve the I-RAVEN task and, differently to the baseline and to our method, they implement an architecture that considers all the nuances of the task (e.g. explicitly modeling panels on the same row). As already mentioned, outperforming specific solutions to the I-RAVEN task is out of scope of our research, rather we want to demonstrate that \emph{sandra} is a competitive generalized reasoner, which is supported by its performance results against the baseline.

\subsection{Domain Generalisation} \label{app:fashion-mnist}
In this section we describe the ontology developed for FashionMNIST (Section \ref{app:fashion-mnist:ontology}) and the architectures used for the experiments (Section \ref{app:fashion-mnist:architectures}.
The code to reproduce the experiments is available at \url{https://anonymous.4open.science/r/sandra-fashionmnist-EEF0/}.
The original Fashion-MNIST dataset provides 10 labels, one per each type of clothes: \emph{T-shirt/top, Trousers, Pullover, Dress, Coat, Sandal, Shirt, Sneaker, Bag, and Ankle boot}. No further semantics is provided about the entities, therefore the ontology developed for FashionMNIST is a straightforward representation modeled after intuitive and common-sense features of the clothes represented in the dataset.

\subsubsection{Fashion-MNIST ontology} \label{app:fashion-mnist:ontology}
We re-engineer the Fashion-MNIST dataset to represent its intensional layer.
The 10 types of clothes are represented as \texttt{rdfs:subClassOf} four top classes: \texttt{:FootWear}, \texttt{:LowerBodyClothes}, \texttt{:UpperBodyClothes}, and \texttt{:Accessories}.
The semantics is introduced in the ontology via several object properties, listed in Table \ref{tab:fmnist-properties}.

For example, the description $\texttt{:Pullover} \in \mathcal{D}$ has roles $\texttt{:LongSleeves}, \texttt{:Neckhole} \in \mathcal{R}$ and it is expressed as a subclass of $\texttt{:UpperBodyClothes} \in \mathcal{D}$.


\subsubsection{Model's architecture} \label{app:fashion-mnist:architectures}
For the domain generalisation experiments two baseline architectures have been used: a straightforward CNN followed by a linear classification head, described in Table \ref{tab:fashion-mnist-architecture-cnn}, and an MLP described in Table \ref{tab:fashion-mnist-architecture-mlp}.

\begin{table}[htb]
    \centering
    \begin{tabular}{l}
        \toprule
        3x3 conv., 16 filters, 2 padding, BN, ReLU \\
        Max pool \\
        3x3 conv., 32 filters, 2 padding, BN, ReLU \\
        Max pool \\
        \textit{Linear layer for projection in \textit{sandra}'s $V$}
        Linear layer \\ \midrule
        Total of $29.0$ K parameters \\
        Total of $84.1$ K parameters with \textit{sandra} \\
        \bottomrule
    \end{tabular}
    \caption{CNN architecture used for the experiments in the Rotated-FashionMNIST dataset.}
    \label{tab:fashion-mnist-architecture-cnn}
\end{table}

\begin{table}[htb]
    \centering
    \begin{tabular}{l}
        \toprule
        Linear layer, $256$ hidden dimension, ReLU \\
        Linear layer, $256$ hidden dimension, ReLU \\
        \textit{Linear layer for projection in \textit{sandra}'s $V$}
        Linear layer \\ \midrule
        Total of $268.7$ K parameters \\
        Total of $278$ K parameters with \textit{sandra} \\
        \bottomrule
    \end{tabular}
    \caption{MLP architecture used for the experiments in the Rotated-FashionMNIST dataset.}
    \label{tab:fashion-mnist-architecture-mlp}
\end{table}

\begin{table*}[ht]
    \centering
    \begin{tabularx}{\textwidth}{XcXX}
        \toprule
        Class & Set & Purpose & Manchester syntax  axiom(s) \\ \midrule
        \texttt{rv:RavenMatrix} & $\mathcal{D}$ & Represents the whole RPM & \texttt{rv:hasPanelSet \textbf{some} rv:PanelSet} \\ \hline
        \texttt{rv:PanelSet} & $\mathcal{D}$ & Sequence of $3$ panels on a row & \texttt{rv:hasPanel \textbf{some} rv:Panel} \\ \hline
        \texttt{rv:Panel} & $\mathcal{D}$ & Represents each of the $9$ panels in an RPM ($16$ if considering the $8$ alternatives to choose from) as shown in Figure \ref{fig:rpm} & \texttt{rv:hasNumberValue \textbf{some} rv:Number}, \texttt{rv:hasComponent \textbf{some} rv:Figure} \\ \hline
        Classes of the form \texttt{rv:PFXY} where $X \in \{ \texttt{rv:A}, \texttt{rv:C}, \texttt{rv:S}, \texttt{rv:T} \}$ and $Y$ depends on $X$. For example, \texttt{rv:A} refers to the angle with $X \in \{0, 1, 2, 3, 4, 5, 6, 7 \}$. & $\mathcal{D}$ & Implements the possible permutations of elements composing and qualifying the panel & \texttt{rv:hasComponent \textbf{some} rv:PXY} \\ \hline
        \texttt{rv:Figure} & $\mathcal{D}$ & Represents each figure in each panel. As shown in Figure \ref{fig:rpm} a figure can vary for several dimensions (color, shape, etc.) & \texttt{rv:hasAngle \textbf{some} rv:Angle}, \texttt{rv:hasColor \textbf{some} rv:Color}, \texttt{rv:hasShape \textbf{some} rv:Shape}, \texttt{rv:hasSize \textbf{some} rv:Size} \\ \hline
        \texttt{rv:Number} & $\mathcal{R}$ & Represents the number of the panel in the RPM. & \\ \hline
        \texttt{rv:Color} & $\mathcal{R}$ & Represents the color of a figure. There are $10$ possible color variations, expressed via $10$ subclasses of the \texttt{rv:Color} class. & \\ \hline
        \texttt{rv:Shape} & $\mathcal{R}$ & Represents the shape of a figure. There are $5$ possible shapes, modeled as subclasses: \texttt{rv:Triangle}, \texttt{rv:Square}, \texttt{rv:Pentagon}, \texttt{rv:Hexagon}, and \texttt{rv:Circle}. & \\ \hline
        \texttt{rv:Angle} & $\mathcal{R}$ & Represents the angle orientation of a figure. There are $8$ possible orientations, expressed as subclasses of the \texttt{rv:Angle} class. & \\ \hline
        \texttt{rv:Size} & $\mathcal{R}$ & Represents the size of a figure. There are $6$ possible sizes, expressed as subclasses of the \texttt{rv:Size} class. & \\
        \bottomrule
    \end{tabularx}
    \caption[]{Detailed description of each class in the I-RAVEN ontology. The \textit{Set} column describes the type of each description, the purpose comments on the class while the axiomatisation is reported using Manchester syntax \cite{horridge2009manchester}.}
    \label{tab:iraven-ontology-classes}
\end{table*}

\begin{table*}[htb]
    \centering
    \begin{tabular}{lccccccccc}
    \toprule
     & \#Params & C & 2$\times$2 & 3$\times$3 & O-IC & O-IG & L-R & U-D  \\ \midrule
     Baseline \cite{zhang2019learning} & 205k & 26.85 & 14.55 & 12.15 & 12.3 & 13.4 & 11.85 & 13.15 \\ 
     WReN \cite{santoro2018raven} & $>$ 1.5M & 29.4 & 26.8 & 23.5 & 22.5 & 21.5 & 21.9 & 21.4 \\
     ResNet \cite{zhang2019learning} & $>$ 25M & 44.7 & 29.3 & 27.9 & 46.2 & 35.8 & 51.2 & 47.4 \\
     ResNet + DRT \cite{zhang2019learning} & $>$ 25M & 46.5 & 28.8 & 27.3 & 46 & 34.2 & 50.1 & 49.8 \\
     LEN \cite{zheng2019raven} & $>$ 5M & 56.4 & 31.7 & 29.7 & 52.1 & 31.7 & 44.1 & 44.2 \\
     CoPINet \cite{chizhang2019raven} & $>$ 5M & 54.4 & 36.8 & 31.9 & 52.2 & 42.8 & 51.9 & 52.5 \\
     DCNet \cite{zhuo2021raven} & $>$ 5M & 57.8 & 34.1 & 35.5 & 57 & 42.9 & 58.5 & 60 \\
     NCD \cite{taozhuo2021raven} & $>$ 11M & 60 & 31.2 & 30 & 62.4 & 39 & 58.9 & 57.2 \\
     SRAN \cite{hu2021stratified} & $>$ 33M & 78.2 & 50.1 & 42.4 & 68.2 & 46.3 & 70.1 & 70.3 \\
     PrAE \cite{zhang2021raven} $\dagger$ & $>$ 150k & 90.5 & 85.35 & 45.60 & 74.7 & 99.7 & 99.5 & 97.4 \\ 
     ConViT \cite{he2023covitraven} & $>$ 5M & 99.9 & 96.2 & 75.5 & 99.4 & 99.6 & 99.5 & 87.3 \\ 
     \cite{hersche2023nesyraven} $\dagger$ & $>$ 11M & 100 & 99.5 & 97.1 & 100 & 100 & 100 & 96.4 \\ \hline
     Baseline + \textit{sandra} & 275k & 45.75 & 16.15 & 14.1 & 14.8 & 14.85 & 13.05 & 13.15 \\
     \bottomrule
    \end{tabular}
    \caption{Results on the accuracy in the I-RAVEN dataset. The number of parameters for related works is a rough estimation based on the architecture used - e.g. we assume that when using a ResNet18 the model will have \textit{at least} $11M$ parameters. It serves as a comparison between the number of parameters of the implemented baseline and those on related works. Results are retrieved from the corresponding papers. Models with $\dagger$ include the use of external solvers, such as planning algorithms.}
    \label{tab:raven-results-app}
\end{table*}

\begin{table*}[ht]
    \centering
    \begin{tabularx}{\textwidth}{XXX}
        \toprule
        F-MNIST property & Purpose & Manchester syntax axiom example  \\ \midrule
        \texttt{:covers} & Expresses which part of the body is covered by a piece of clothing, if any & \texttt{:Dress :covers \textbf{some} :WholeBody}, \texttt{:Sandal :covers \textbf{some} :Feet} \\ \hline
        \texttt{:hasClosure} & Expresses the type of closure of the piece of clothing (e.g. zip, buttons, laces, etc.), if any & \texttt{:Coat :hasClosure \textbf{some} :ButtonClosure} \\ \hline
        \texttt{:hasShape} & Expresses if the entity covers entirely the body part or if it leaves some uncovered parts & \texttt{:Sandal :hasShape \textbf{some} :OpenShape}, \texttt{:Sneaker :hasShape \textbf{some} :ClosedShape} \\ \hline
        \texttt{:hasSleeves} & Expresses the entity's kind of sleeves, if any  & \texttt{:Pullover :hasSleeves \textbf{some} :LongSleeves}, \texttt{:T-shirt\_top :hasSleeves \textbf{some} :ShortSleeves}  \\ \hline
        \texttt{:wearingContext} & Expresses the prototypical context in which a certain entity can be worn  & \texttt{:AnkleBoot :wearingContext \textbf{some} :FormalContext}, \texttt{:T-shirt\_top :wearingContext \textbf{some} :InformalContext}  \\ \hline
        \texttt{:wearingPoint} & Expresses the part of the body from which an item of clothing is first worn & \texttt{:Coat :wearingPoint \textbf{some} :SleevesHole}, \texttt{:Pullover :wearingPoint \textbf{some} :NeckHole}  \\
        \bottomrule
    \end{tabularx}
    \caption[]{Details of each property used to axiomatize each description in the ontology. Axioms examples can be seen in the Axiomatization column, written in Manchester syntax \cite{horridge2009manchester}.}
    \label{tab:fmnist-properties}
\end{table*}

\end{document}